\documentclass[11pt]{article}
\pdfoutput=1
\usepackage{acl}

\usepackage{times}
\usepackage{latexsym}

\usepackage[T1]{fontenc}

\usepackage[utf8]{inputenc}

\usepackage{microtype}

\usepackage{todonotes}
\usepackage{float}
\usepackage{subcaption}
\usepackage{lipsum}
\usepackage{graphicx}

\usepackage{amssymb}
\usepackage{amsthm}
\usepackage{amsmath}

\usepackage{multirow}
\usepackage{booktabs}

\usepackage{algpseudocode}
\usepackage{algorithm}

\usepackage{longtable} 
\usepackage{supertabular}

\title{Principled Paraphrase Generation with Parallel Corpora}

\author{Aitor Ormazabal$^{1}$, \ Mikel Artetxe$^{2}$, \ Aitor Soroa$^{1}$, \ Gorka Labaka$^{1}$, \ Eneko Agirre$^{1}$ \\
$^1$HiTZ Center, University of the Basque Country (UPV/EHU) \\
$^2$Meta AI \\
\texttt{\{aitor.ormazabal,a.soroa,gorka.labaka,e.agirre\}@ehu.eus} \\
\texttt{artetxe@fb.com}
}

\date{}

\newtheorem{theorem}{Theorem}

\begin{document}

\maketitle

\begin{abstract}
Round-trip Machine Translation (MT) is a popular choice for paraphrase generation, which leverages readily available parallel corpora for supervision. In this paper, we formalize the implicit similarity function induced by this approach, and show that it is susceptible to non-paraphrase pairs sharing a single ambiguous translation.
Based on these insights, we design an alternative similarity metric that mitigates this issue by requiring the entire translation distribution to match, and implement a relaxation of it through the Information Bottleneck method. Our approach incorporates an adversarial term into MT training in order to learn representations that encode as much information about the reference translation as possible, while keeping as little information about the input as possible. Paraphrases can be generated by decoding back to the source from this representation, without having to generate pivot translations. In addition to being more principled and efficient than round-trip MT, our approach offers an adjustable parameter to control the fidelity-diversity trade-off, and obtains better results in our experiments.

\end{abstract}
\section{Introduction}

Paraphrase generation aims to generate alternative surface forms expressing the same semantic content as the original text~\cite{madnani2010generating}, with applications in language understanding and data augmentation~\cite{zhou-bhat-2021-paraphrase}. 
One popular approach is to use an MT system to translate the input text into a pivot language and back  \citep{wieting-gimpel-2018-paranmt,mallinson-etal-2017-paraphrasing,  roy-grangier-2019-unsupervised}. While it intuitively makes sense that translating to another language and back should keep the meaning of a sentence intact while changing its surface form, it is not clear what exactly would be considered a paraphrase by such a system.

\begin{figure}
    \centering
    \includegraphics[width=\columnwidth]{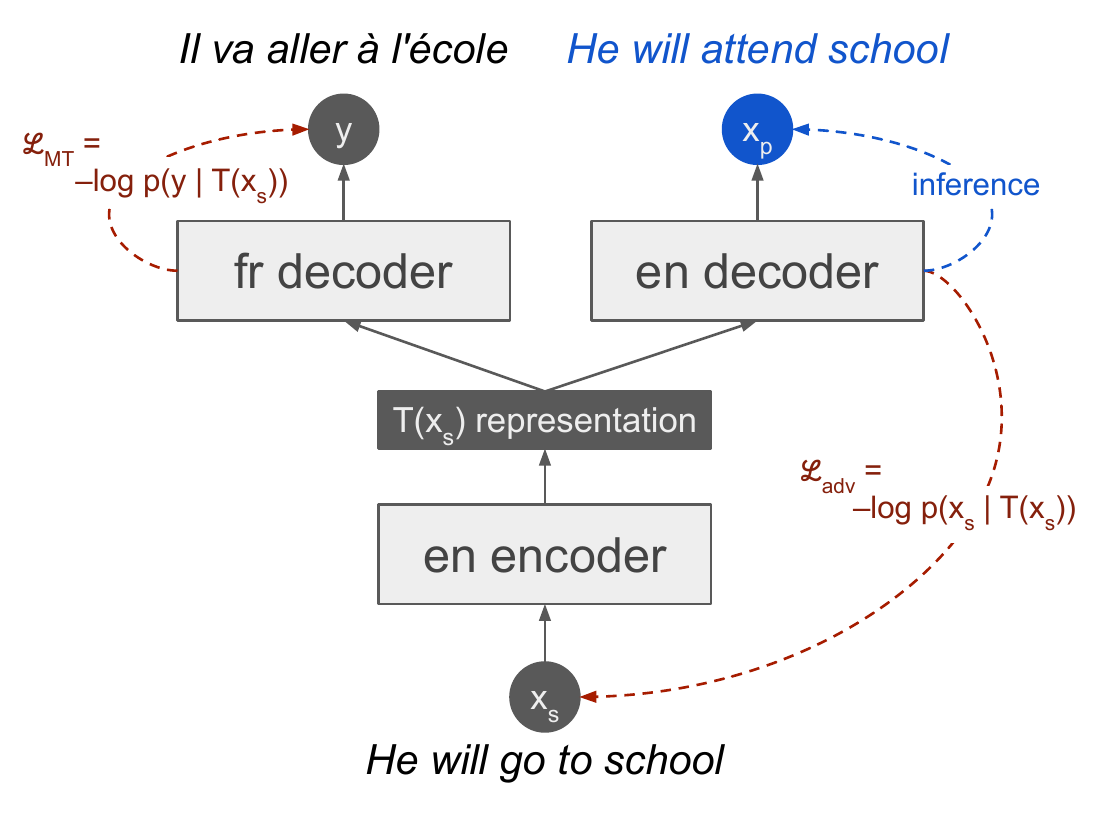}
    \caption{\textbf{Proposed system.}
    Given the input $x_s$, we aim to learn a representation $T(x_s)$ that encodes as much information as possible about it's reference translation $y$ (ensuring that the meaning is preserved), and as little information as possible about $x_s$ itself (ensuring that surface information is removed). We achieve this through adversarial learning, where the encoder minimizes $\lambda \mathcal{L}_{MT} - (1-\lambda) \mathcal{L}_{adv}$ and the two decoders minimize $\mathcal{L}_{MT}$ and $\mathcal{L}_{adv}$. At inference time, we couple the English encoder and decoder to generate a paraphrase $x_p$ which, being conditioned on $T(x)$, will preserve the meaning of $x_s$ but use a different surface form.}

    \label{fig:system}
\end{figure}

\begin{figure*}
    \centering
    \includegraphics[width=0.8\textwidth]{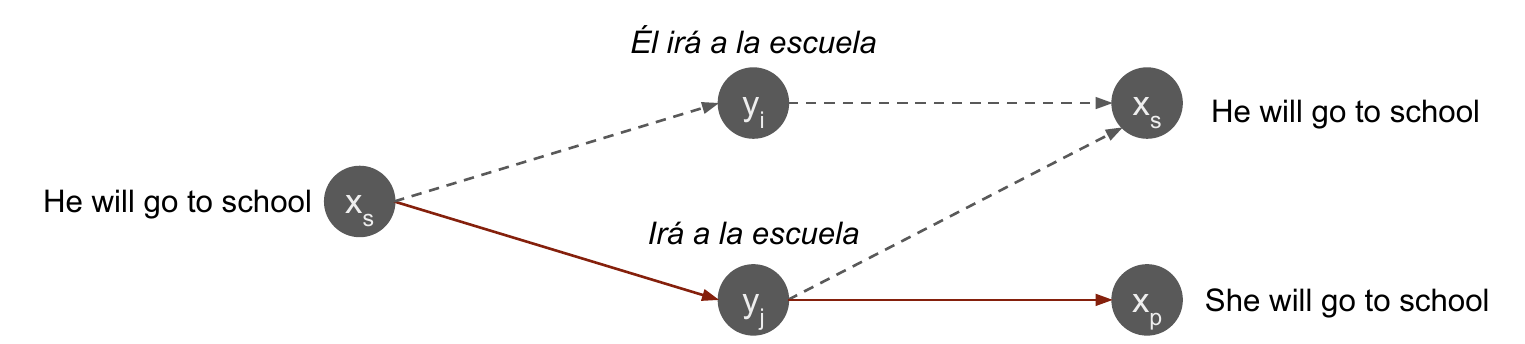}
    \caption{\textbf{Confounding translation problem in round-trip MT.} \textit{``Irá a la escuela''} does not mark the gender of the subject due to ellipsis, and it is thus a valid translation of both \textit{``He will go to school''} and \textit{``She will go to school''}. As a consequence, round-trip MT could generate \textit{``She will go to school''} as a paraphrase of \textit{``He will go to school''}. Our approach mitigates this issue by requiring the full translation distribution to match.}
    \label{fig:mtissue}
\end{figure*}

In this work, we show that the probability of a paraphrase $x_p$ given a source sentence $x_s$ under a round-trip MT system can be naturally decomposed as $P(x_p|x_s) = P(x_p)S(x_p,x_s)$, where $S$ is a symmetric similarity metric over the paraphrase space and $P(x_p)$ the probability of $x_p$. We argue that this similarity function is not appropriate in the general case, as it can assign a high score to sentence pairs that share an ambiguous translation despite not being paraphrases of each other. This phenomenon is illustrated in Figure \ref{fig:mtissue}, where $x_s$ and $x_p$ share a confounding translation without gender marker. %

So as to address this issue, we design an alternative similarity function that requires the entire translation distribution to match, and develop a relaxation of it through the Information Bottleneck (IB) method. We implement this approach using an adversarial learning system depicted in Figure \ref{fig:system}. Our model combines an encoder that, for a given sentence, removes the information that is not relevant to predict its translation, and a decoder that reconstructs a paraphrase from this encoding. In addition to being more principled, our approach is more efficient than round-trip MT at inference, can be tuned to favor fidelity or diversity, and achieves a better trade-off between the two. Our code is freely available \footnote{\url{https://github.com/aitorormazabal/paraphrasing-from-parallel}}.

\section{Related Work}

We next review the paraphrase generation literature (\S\ref{subsec:paraphrasing}), and describe the information bottleneck method (\S \ref{subsec:ib}), which is the basis of our proposal.

\subsection{Paraphrase generation}
\label{subsec:paraphrasing}

Early work on paraphrasing focused on retrieval methods, either extracting plausible sentences from large corpora for generation \citep{barzilayextracting, bannardparaphrasing}, or identifying paraphrase pairs from weakly aligned corpora to create paraphrase datasets \citep{coster-kauchak-2011-simple,dolan-etal-2004-unsupervised}. More recently, neural approaches for paraphrase generation have dominated the field. We classify these methods according to the type of supervision they use.

\paragraph{Monolingual corpora.} These systems are trained in an unsupervised fashion using unlabeled monolingual corpora. They usually employ an information bottleneck, with the goal of encoding semantic information in the latent space. Approaches include Variational Autoencoders (VAE) \citep{bowman-etal-2016-generating}, VAEs with Vector Quantization \citep{roy-grangier-2019-unsupervised}, and latent bag-of-words models \citep{yaobowlatent}.  \citet{huang-chang-2021-generating} disentangle semantic and syntactic content in the latent space through a bag of words representation, which allows for syntactically controllable generation.%

\paragraph{Parallel corpora.} These systems are trained on pairs of parallel sentences in two languages. Most of these methods are based on round-trip MT, where a sentence is translated to a pivot language and back in order to obtain a paraphrase. \citet{huparabank} add lexical constraints to the MT decoding procedure to obtain better paraphrases. \citet{mallinson-etal-2017-paraphrasing} generate not one but multiple pivot sentences and use a fusion-in-decoder strategy.

\paragraph{Paraphrase corpora.} These systems are trained in a supervised manner over pairs or clusters of paraphrases. When such data is available, training a regular sequence-to-sequence model is a strong baseline \citep{egonmwan-chali-2019-transformer-seq2seq}. \citet{kumar-etal-2019-submodular} add submodular optimization to improve paraphrase diversity.
Some VAE-based methods also leverage paraphrase clusters to learn a latent representation that disentangles meaning and form \citep{iyyer-etal-2018-adversarial, 10.1162/tacl_a_00318, hosking2021factorising, chen-etal-2019-controllable}. Most of these methods require a syntactic exemplar for generation, and assume that all surface forms are valid for all sentences. \citet{hosking2021factorising} do away with this assumption in the context of question paraphrasing, predicting a valid  syntactic embedding from a discrete set at test time. %

\medskip

While it is paraphrase corpora that offers the strongest supervision, such data is hard to obtain and usually restricted to narrow domains like Quora Question Pairs, WikiAnswers and Twitter \citep{hosking2021factorising, kumar-etal-2019-submodular, egonmwan-chali-2019-transformer-seq2seq}. In contrast, parallel corpora is widely available, while offering a stronger training signal than monolingual corpora. For that reason, round-trip MT is a common choice when paraphrases are needed for downstream tasks \citep{xie-etal-unsupervised,artetxe-etal-2020-translation}, as well as a common baseline in the paraphrasing literature \citep{hosking2021factorising,roy-grangier-2019-unsupervised}.\footnote{Round-trip MT has also been used to generate synthetic paraphrase corpora \citep{wieting-gimpel-2018-paranmt}.} Our work focuses on this class of systems, identifying the limitations of round-trip MT and proposing a more principled alternative.

\subsection{The Information Bottleneck Method} \label{subsec:ib}

Given two random variables $X,Y$, the Information Bottleneck (IB) method \citep{Tishby99theinformation} seeks to learn a representation $T(X)$ that minimizes the Mutual Information (MI) between $T$ and $X$, while preserving a minimum MI between $T$ and $Y$. That is, the objective $I(X,T) \; s.t.\; I(T,Y)\geq \gamma $ is minimized. Since the MI is usually impossible to calculate exactly for neural representations, a common approach is to use variational methods, that turn the estimation problem into an optimization one. This can be done by adding a neural decoder on top of the representation, and training the entire system end-to-end \citep{pmlr-v97-poole19a}. This is the approach we follow in this work.

\section{Characterizing Round-trip MT}

\label{sec:impfunc}

Let $X$ be a random variable representing a sequence in the source language, and $Y$ be a random variable representing its translation into a pivot language.\footnote{For convenience, we will also use $X$ and $Y$ to refer to the set of source and target language sequences, and abbreviate probabilities of the form $P(X=x)$ as $P(x)$.}
Given an input sequence $x_s \in X$, we can use round-trip MT to generate a paraphrase $x_p \in X$ by translating $x_s$ into the pivot language and back, according to the forward and backward translation models $P(y|x_s)$ and $P(x_p|y)$. As such, we can formulate the probability of round-trip MT generating a particular paraphrase $x_p$ by marginalizing over the set of possible pivot translations: 
\begin{equation}
P(x_p | x_s) = \sum_{y \in Y} P(y | x_s) P(x_p|y)
\label{eq:mt_original}
\end{equation}
In what follows, we will characterize the paraphrases produced by this approach, i.e. the properties that $x_p$ needs to meet in relation to $x_s$ for $P(x_p | x_s)$ to be high.\footnote{Some round-trip MT systems do not consider all possible translations into the pivot language, but only a subset of them \citep{mallinson-etal-2017-paraphrasing}. In that case, the sum in Eq. \ref{eq:mt_original} goes over $y \in \{y_1, ..., y_K\}$, and we need to introduce a partition $Z=  \sum_{y \in \{y_1,...,y_K\} } P(y | x_s)$ to normalize the probabilities. However, the fundamental analysis in this section still applies. Refer to Appendix \ref{app:topk} for more details.}

By applying Bayes' rule, we can rewrite Eq. \ref{eq:mt_original} as follows:
\begin{equation}
P(x_p | x_s) = P(x_p) \underbrace{\sum_{y \in Y} \frac{ P(y | x_s) P(y|x_p)}{P(y)}}_{S_{MT}(x_p, x_s)}
\label{eq:mtdecomp}
\end{equation}
The sum on the right hand side can be interpreted as a symmetric similarity function, $S_{MT}(x_p, x_s) = S_{MT}(x_s, x_p) = \sum_y \frac{ P(y | x_s) P(y|x_p)}{P(y)}$, which measures the likelihood of two sentences to be actual paraphrases. The probability of $x_p$ given $x_s$ then becomes $P(x_p | x_s) = P(x_p)S_{MT}(x_p,x_s)$, which is the similarity between $x_s$ and $x_p$, weighted by the marginal probability of $x_p$. 

But when are $x_s$ and $x_p$ considered similar according to the above definition of $S_{MT}(x_s, x_p)$? Intuitively, $S_{MT}$ is a measure of the \textit{overlap} between the conditional distributions that $x_s$ and $x_p$ induce over $Y$. This will be highest when $P(y | x_s) P(y|x_p)$ is as large as possible for as many $y$ as possible. At the same time, $P(y | x_s) P(y|x_p)$ will be high when both $P(y | x_s)$ and $P(y | x_p)$ are high, that is, when $y$ is a probable translation of both $x_s$ and $x_p$. This captures the intuition that two sentences are similar when they can be translated into the same text in the pivot language.

But what if $x_s$ and $x_p$ have one particular high-probability translation $y_j$ in common, but differ in the rest? As illustrated in Figure \ref{fig:mtissue}, this can happen when $y_j$ is ambiguous in the target language and can mean both $x_s$ and $x_p$, even if $x_s$ and $x_p$ are not equivalent (e.g., when $x_s$ uses the masculine form, $x_p$ the feminine form, and $y_j$ does not mark the gender). In this case, the sum $\sum_y \frac{ P(y | x_s) P(y|x_p)}{P(y)}$ will be dominated by $\frac{ P(y_j | x_s) P(y_j|x_p)}{P(y_j)}$, which will be high when both $P(y_j | x_s)$ and $P(y_j|x_p)$ are high.

We can thus conclude that the implicit similarity function underlying round-trip MT is flawed, as it assigns a high score to a pair of sequences $(x_s, x_p)$ that have an ambiguous translation in common. As a consequence, round-trip MT will generate $x_p$ as a paraphrase of $x_s$ with a high probability, even if the two sequences have a different meaning.

\section{Principled Paraphrasing}
\label{sec:principled}

As shown in the previous section, the implicit similarity function induced by round-trip MT is not adequate in the general case, as it assigns a high score to pairs of sequences that share a single translation, despite differing in the rest. So as to address this, we can define an alternative similarity function that requires the entire translation distribution to match: 
\begin{equation}
 \label{eq:sim}
     S(x_p, x_s) = 
     \begin{cases}
      1 & 
      P(y|x_p) = P(y|x_s) \forall y\in Y \\
      0 & \text{otherwise}
     \end{cases}
\end{equation}
and use it to replace $S_{MT}$ in Eq. \ref{eq:mtdecomp} so that $P(x_p|x_s) \propto P(x_p)S(x_p, x_s)$.
However, this definition is too strict, as it is virtually impossible that $P(y|x_p)$ and $P(y|x_s)$ are exactly the same for all $y \in Y$.\footnote{One reason is that we use empirical estimates of $P(y|x_p)$ and $P(y|x_s)$, which will deviate from the ground truth.} In \ref{subsec:ib-implementation}, we define a relaxation of it through the IB method, which introduces an adjustable parameter $\beta$ to control how much we deviate from it. In \ref{subsec:ib-characterization}, we characterize the paraphrases generated by this approach, showing that they are less susceptible to the problem of confounding translations described in the previous section.

\subsection{IB-based relaxation}
\label{subsec:ib-implementation}

So as to implement the similarity function in Eq. \ref{eq:sim}, we will use the IB method to learn an encoding $T$ for $X$ such that the following holds:
\begin{equation}
 S(x_p, x_s)  =  \frac{P(x_p | T(x_s))}{P(x_p) Z(x_s)}
 \end{equation}
where $Z(x_s)$ is a normalizer that does not depend on the paraphrase candidate $x_p$. %

As seen in \S\ref{subsec:ib}, given a source variable $X$ and a target variable $Y$, the IB method seeks to find an encoding $T(X)$ that minimizes the MI with $X$ (maximizing compression), while preserving a certain amount of information about $Y$:
 \begin{equation}
 \min_T I(X,T) \; s.t \; I(T,Y) \geq \gamma.
 \end{equation}
This constrained minimization is achieved by introducing a Lagrange multiplier $\beta$ and minimizing
 \begin{equation}
 \min_T I(X,T) - \beta I(T,Y).
 \end{equation}
As $\beta \rightarrow \infty $, all the information about $Y$ is preserved and the IB method learns a minimal sufficient statistic $T$, that is, an encoding that satisfies $I(T,Y) = I(X,Y)$ while achieving the lowest $I(T,X)$ possible. The following theorem states that such a minimal sufficient statistic $T$ induces the similarity function in Eq. \ref{eq:sim} (proof in Appendix~\ref{app:proofs}):
 \begin{theorem}
 \label{theorem:IB}
Suppose the random variable X represents a sentence in the source language, Y represents its translation, and T is a minimal sufficient statistic of X with respect to Y. Let $x_p$ and $x_s$ be a pair of sentences in the source language. Then, $P(x_p| T(x_s)) = P(x_p)\frac{S(x_p, x_s)}{Z(x_s)}  $, where $S$ is given by Equation \ref{eq:sim}, and $Z$ is a normalizing factor that does not depend on $x_p$.
\end{theorem}

Thus, as $\beta \rightarrow \infty$ the IB method approximates the similarity metric $S$. In practice, when $\beta$ is set to a fixed finite number, losing some information about the target variable is allowed, and a relaxation of the metric $S$ is learned instead. %

\subsection{Characterizing IB-based paraphrasing}
\label{subsec:ib-characterization}

We will next analyze the relaxation of $S$ induced by the IB method. We will characterize what kind of sentences are considered paraphrases by it, showing that it is less susceptible to the problem of confounding translations found in round-trip MT (\S\ref{sec:impfunc}). Derivations for the results in this section, as well as alternative bounds and broader discussion can be found in Appendix \ref{sec:charac}.

As seen in \S\ref{subsec:ib-implementation}, we define paraphrase probabilities given an encoding T as $P(x_p| T(x_s)) =P(X=x_p | T(X) = T(x_s))$, which can only be non-zero if $T(x_p)=T(x_s)$. This means that the encoding $T$ will partition the source space into a collection of paraphrase clusters according to its value. Mathematically, given the equivalence relation $x_1\sim x_2 \iff T(x_1)=T(x_2)$, only sentence pairs within the same equivalence class will have non-zero paraphrase probabilities. We then have the following theorem:
\begin{theorem}
\label{theorem:7}
Suppose $T$ is a solution of the IB optimization problem $\min_T I(X,T) \; s.t \; I(T,Y) \geq \gamma$, and $\epsilon = I(X,Y) - \gamma$. If $\mathcal{A}$ is the partition on $X$ induced by $T$, we have: 
\begin{equation}
\begin{split}
\sum_{A\in \mathcal{A}} \max_{x_1, x_2\in A} \frac{P(x_1)P(x_2)}{2(P(x_1) + P(x_2))} \\\cdot D_1(P_{Y|x_1}, P_{Y|x_2})^2 \leq \epsilon,
\end{split}
\end{equation}
where $D_1$ is the $L_1$ norm distance. 
\end{theorem}

It is easy to see that, when $\epsilon = 0$, corresponding to $\gamma = I(X,Y)$ and $\beta\rightarrow \infty$, this forces all distances to be zero. In that case, only sentences with identical translation distributions are considered paraphrases, in accordance with Theorem \ref{theorem:IB}.

In the general case, Theorem %
\ref{theorem:7} 
states that the $L_1$ distance between the translation distributions of sentences that are considered paraphrases cannot be high, as it will be bounded by a function of $\epsilon$.
While the $S_{MT}$ metric in $\S\ref{sec:impfunc}$ can be dominated by a high-probability term and effectively ignore differences in probability for the less likely translations, the $L_1$ norm gives equal importance to differences in probability for every translation candidate. 
Thanks to this, the resulting system will be less susceptible to the problem of confounding translations.

\section{Proposed System}
\label{sec:system}

In this section, we describe a practical implementation of the IB-based paraphrasing approach defined theoretically in \S\ref{sec:principled}.

As illustrated in Figure \ref{fig:system}, our system can be seen as an extension of a regular encoder-decoder MT architecture with an additional adversarial decoder, which is trained with an auto-encoding objective to reconstruct the original sentence $x_s$ from the encoder representation $T(x_s)$. The encoder is trained to minimize the cross-entropy loss of the MT decoder, while maximizing the loss of the adversarial decoder. This way, the encoder is encouraged to remove as much information about $x_s$ as possible, while retaining the information that is necessary to predict its reference translation $y$. Thanks to this, $T(x_s)$ should capture the semantic content of $x_s$ (which is relevant to predict $y$), without storing additional surface information (which is not relevant to predict $y$). Once the model is trained, the adversarial decoder can be used to generate paraphrases of $x_s$ from this representation $T(x_s)$. %

This adversarial architecture can be interpreted as an implementation of the IB method as follows. Following \citet{pmlr-v97-poole19a}, we start by adding a decoder $q(y|t)$ on top of the encoder $T(x)$, and rewrite the $I(T,Y)$ term as:
\begin{equation}
\label{eq:beginsplit-it-y}  
\begin{split}
    I(T,Y) &= \mathbb{E}_{P(y,t)} \bigg[ \log \frac{q(y|t)}{P(y)} \bigg] \\
    &+ \mathbb{E}_{P(t)}[KL(P(y|t)||q(y|t))] \\
    &\geq \mathbb{E}_{P(y,t)} \bigg[ \log q(y|t) \bigg]  + h(Y), 
\end{split}
\end{equation}
where equality will hold if $q$ is the true conditional distribution $q(y|t) = P(y|t)$, and $h$ is the differential entropy. If we parametrize $T$ and $q$ by a neural network encoder-decoder architecture the $\mathbb{E}_{P(y,t)} \bigg[ \log q(y|t) \bigg]$ term in Eq. \ref{eq:beginsplit-it-y}, can be rewritten as $\mathbb{E}_{P(y,x)} \bigg[ \log q(y|T(x)) \bigg]$, which is precisely the log likelihood of the data distribution of $X,Y$ given by $P$. In other words, by training the encoder-decoder to maximize Eq. \ref{eq:beginsplit-it-y}, we are implicitly maximizing the mutual information $I(T,Y)$.

Similarly, one can approximate
\begin{equation}
\begin{split}
    I(X,T) &\geq \mathbb{E}_{P(x,t)} \bigg[ \log q(x|t) \bigg]  + h(X) 
    \\ &=  \mathbb{E}_{P(x)} \bigg[ \log q(x|T(x)) \bigg]  + h(X) ,
\end{split}
\end{equation}
where equality will hold when $q$ is the true conditional distribution and  $q(x|T(x)) = P(x|T(x))$. Thus, given an ideal decoder $q$ that perfectly approximates the conditional distributions $q(x|T(x))$ and $q(y|T(x))$, the IB minimization problem is equivalent to minimizing
\begin{equation}
\label{eq:nll_loss_orig}
\begin{split}
  \mathbb{E}_{p(x)} \bigg[ \log q(x|T(x)) \bigg]  - \beta \mathbb{E}_{P(y,t)} \bigg[ \log q(y|t) \bigg] 
  \\=   \mathbb{E}_{P(x,y)} \bigg[ \log q(x|T(x))  - \beta  \log q(y|T(x)) \bigg]. 
\end{split}
\end{equation}
In practice, we parametrize both the encoder $T$ and the decoder $q$ with transformer neural networks, and learn them from a parallel corpus. Since $\log q(y|T(x)) $ is a lower bound of $I(T,Y) - h(Y)$, maximizing this term is theoretically sound.  Minimizing $\mathbb{E}_{P(x)} \bigg[ \log q(x|T(x)) \bigg] $, on the other hand, amounts to minimizing a lower bound, which, while not as theoretically solid, is common practice in the variational optimization literature \citep{NEURIPS2018_1ee3dfcd, pmlr-v80-kim18b}. 

Finally, we reparametrize Eq. \ref{eq:nll_loss_orig} by setting $\lambda = \frac{\beta}{1+\beta}$ to obtain the equivalent minimization objective
\begin{equation}
\label{eq:nll_loss_final}
\begin{split}
  \mathcal{L}(T,q) = \mathbb{E}_{P(x,y)} [  - \lambda  \log q(y|T(x)) \\ + (1-\lambda) \log q(x|T(x))] = \\\lambda \mathcal{L}_{MT}(T,q) - (1-\lambda) \mathcal{L}_{Adv}(T), 
\end{split}
\end{equation}
where $\mathcal{L}_{MT}$ is the regular MT loss of cross-entropy with the translation target, and $\mathcal{L}_{Adv}$ is the cross-entropy with the source sentence (see Figure \ref{fig:system}).\footnote{We make the adversarial term a function of $T$ only in the minimization objective, as the gradient from the adversarial term is only propagated to the encoder. The adversarial decoder is independently trained to predict the source  from the encoded representation.} We thus observe that the proposed adversarial training architecture approximates the IB method. The setting $\beta \rightarrow \infty$ corresponds to $\lambda \rightarrow 1$,  where the optimal solution is a minimal sufficient statistic. 

During training, the expectation in Eq. \ref{eq:nll_loss_final} is approximated by sampling batches from the training data. Care must be taken when optimizing the loss, as we do not want to propagate gradients of the adversarial loss to the adversarial decoder. If we did, a trivial way to minimize $(1-\lambda) \log q(x|T(x))$ would be to make the decoder bad at recovering $x$, which would not encourage $T(x)$ to encode as little information as possible. To prevent this, we use a percentage $K$ of the batches to learn the adversarial decoder $\log q(x|T(x))$, where the encoder is kept frozen. The rest of the batches are used to optimize the full term $ - \lambda  \log q(y|T(x))  + (1-\lambda) \log q(x|T(x)) $, but the gradients for the second term are only propagated to the encoder.

\section{Experimental Design}

We experiment with the following systems:
\begin{itemize}
    \item \textbf{Proposed.} Our system described in \S\ref{sec:system}. We share the weights between the MT decoder and the adversarial decoder, indicating the language that should be decoded through a special language ID token. Unless otherwise indicated, we use $\lambda = 0.73 $ and $K = 0.7$, which performed best in the development set.\footnote{We performed a grid search, where $\lambda \in \{0.7, 0.73, 0.8\}$ and $K \in \{ 0.7, 0.8 \}$, and chose the checkpoint with best iBLEU with $\alpha = 0.7$.}
    \item \textbf{Round-trip MT.} A baseline that uses two separate MT models to translate into a pivot language and back (see \S\ref{sec:impfunc}).%
    \item \textbf{Copy.} A baseline that copies the input text.
\end{itemize}
We use mBART \cite{liu2020multilingual} to initialize both our proposed system and round-trip MT, and train them using the same hyper-parameters as in the original work.\footnote{$0.3$ dropout, $0.2$ label smoothing, $2500$ warm-up steps, $3e-5$ maximum learning rate, and $100K$ total steps.} In both cases, we use the English-French WMT14 dataset~\cite{bojar2014findings} as our parallel corpus for training.\footnote{We filter the dataset by removing sentence pairs with a source/target length ratio that exceeds $1.5$ or are longer than 250 words.} We report results for two decoding strategies: beam search with a beam size of 5, and top-10 sampling with a temperature of 0.9 (optimized in the development set).\footnote{In the case of round-trip MT, we always use beam search to generate the pivot translation, and compare the two approaches to generate paraphrases from it.}

We consider two axes when evaluating paraphrases: \textit{fidelity} (the extent to which the meaning of the input text is preserved) and \textit{diversity} (the extent to which the surface form is changed). Following common practice, we use a corpus of gold paraphrases to automatically measure these. More concretely, given the source sentence $s$, the reference paraphrase $r$ and the candidate paraphrase $c$, we use BLEU($c, r$) as a measure of fidelity, and BLEU($c, s$)---known as self-BLEU---as a measure of diversity. An ideal paraphrase system would give us a high BLEU, with as low a self-BLEU as possible. Given that there is generally a tradeoff between the two, we also report iBLEU = $\alpha$ BLEU $- (1-\alpha)$ self-BLEU, which combines both metrics into a single score \citep{mallinson-etal-2017-paraphrasing}. Following \citet{hosking2021factorising}, we set $\alpha = 0.7$.

For development, we extracted 156 paraphrase pairs from the STS Benchmark dataset \cite{cer-etal-2017-semeval}, taking sentence pairs with a similarity score above 4.5. For our final evaluation, we used the Multiple Translations Chinese (MTC) corpus \citep{huang2002mtc}, which comprises three sources of Chinese journalistic text translated into English by multiple translation agencies. We extract the translations of the first two agencies to obtain an test set of 993 paraphrase pairs, where one is the source and the other the reference paraphrase. The third sentence if kept as an additional paraphrase for estimating human performance.

\begin{table}
\begin{center}
\addtolength{\tabcolsep}{-1.5pt}
\resizebox{1\linewidth}{!}{
\begin{tabular}{l|c|c|c}
               & Self-BLEU $\downarrow$ & BLEU $\uparrow$ & iBLEU $\uparrow$ \\ 
Model          & (diversity) & (fidelity) & (combined)  \\ \hline \hline
Copy           & 100.0       & 23.0 & -13.9 \\ %
MT (beam)       & 51.1      & 18.8 & -2.17 \\ 
MT (sampling)  & 41.4 & 15.8 & -1.36 \\ \hline
Ours (beam)     & 33.0      & 15.5 & 0.95  \\ 
Ours (sampling) & 27.3      & 13.2 & \textbf{1.05}  \\ \hline
Human   &  18.1    & 19.8  & 8.43 \\ \hline
\end{tabular}
}
\end{center}
\caption{\textbf{Results on the MTC dataset for three baselines (top rows), our two systems, and human performance.} $\downarrow$ smaller is better, $\uparrow$ larger is better. }
\label{tab:automatic}
\end{table}

\section{Results}

We next report our main results (\S\ref{sec:automaticeval}), followed by a qualitative analysis (\S\ref{sec:humaneval}).

\subsection{Main results}
\label{sec:automaticeval}

We report our main results in Table \ref{tab:automatic}. As it can be seen, our proposed system outperforms all baselines in terms of iBLEU, indicating that it achieves a better trade-off between diversity and fidelity. This advantage comes from a large improvement in diversity as measured by self-BLEU, at a cost of a small drop in fidelity as measured by BLEU. Both for round-trip MT and our proposed system, beam search does better than sampling in terms of fidelity, at the cost of sacrificing in diversity. Finally, the human reference scores show ample room for improvement in both axes.

\begin{figure}
    \centering
    \includegraphics[width=\columnwidth]{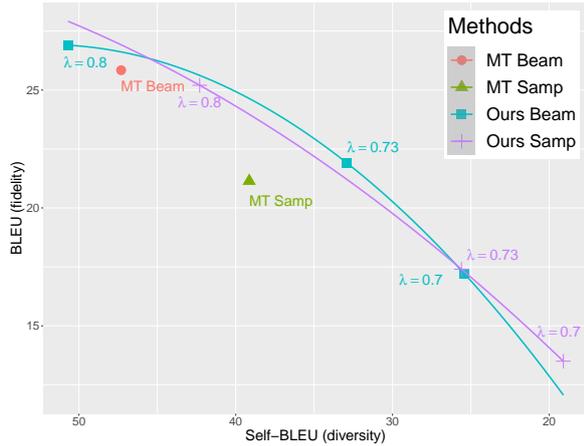}
    \caption{\textbf{Effect of varying the $\lambda$ parameter on the development set.} BLEU in the vertical axis. The horizontal self-BLEU axis is mirrored, so systems toward the top right have the best trade-off between diversity and fidelity.}
    \label{fig:lambdachart}
\end{figure}

While our proposed system achieves the best combined score, our results also show that different approaches behave differently in terms of diversity and fidelity. In practice, it would be desirable to have a knob to control the trade-off between the two, as one may want to favor diversity or fidelity depending on the application. One additional advantage of our approach over round-trip MT is that it offers an adjustable parameter $\lambda$ to control the trade-off between these two axes. So as to understand the effect of this parameter, we tried different values of it in the development set, and report the resulting curve in Figure \ref{fig:lambdachart} together with the MT baselines. BLEU and Self-BLEU scores of the best checkpoints for each $\lambda$ (0.7,0.73,0.8) and plot the results together with the MT baselines for our systems in Figure \ref{fig:lambdachart}.

As expected, higher values of $\lambda$ yield systems that tend to copy more, being more faithful but less diverse. Consistent with our test results, we find that, for a given value of $\lambda$, beam search does better than sampling in terms of fidelity, but worse in terms of diversity, yet
both decoding strategies can be adjusted to achieve a similar trade-off. More importantly, we observe that both curves are above round-trip MT, the gap being largest for the sampling variant. We can thus conclude that our proposed approach does better than round-trip MT for a comparable trade-off between diversity and fidelity, while offering a knob to adjust this trade-off as desired. %

\begin{figure}
    \centering
    \includegraphics[width=\columnwidth]{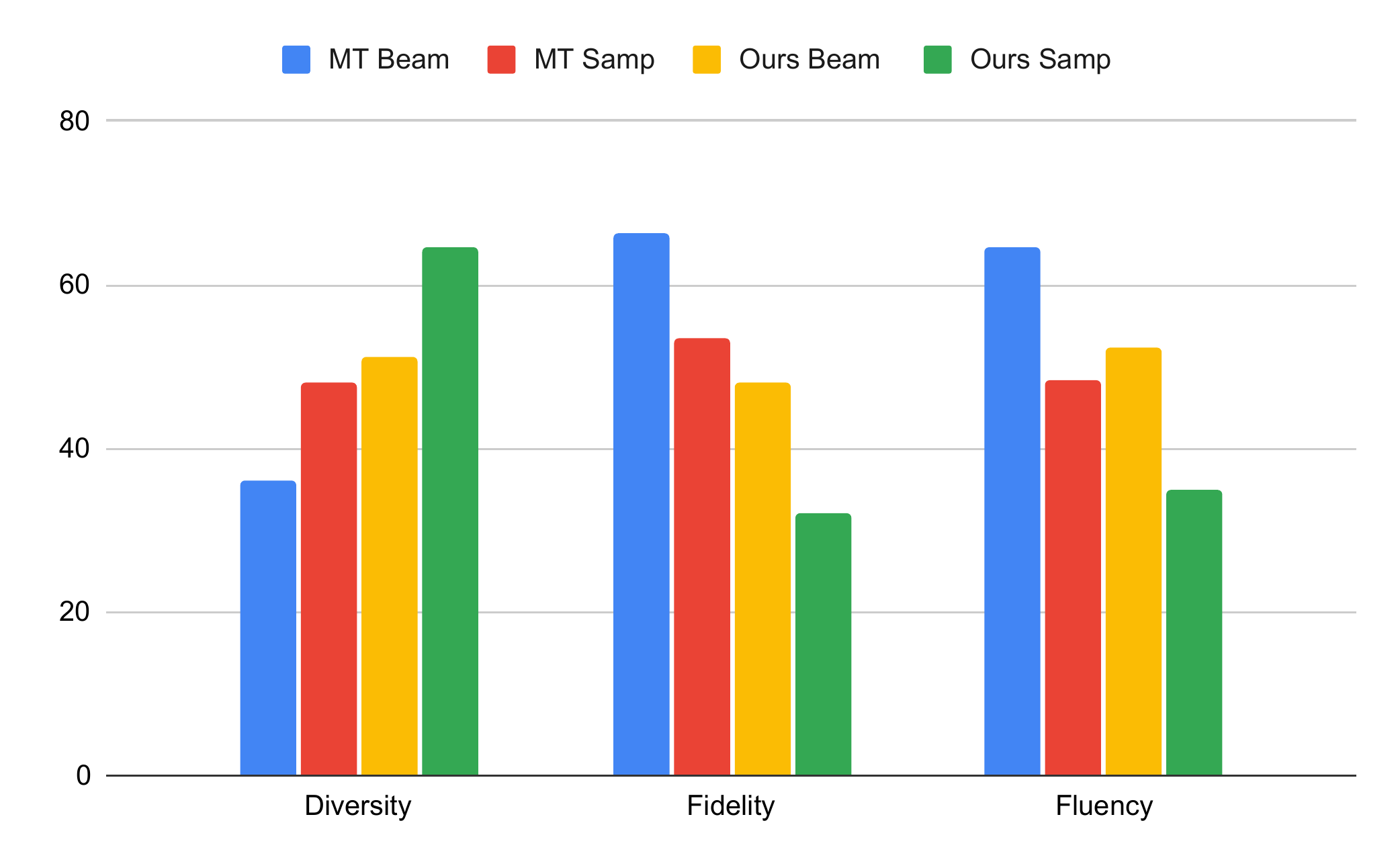}
    \caption{\textbf{Human evaluation results (larger is better).} Refer to \S\ref{sec:humaneval} for more details.}
    \label{fig:human}
\end{figure}

\begin{table*}[ht]
\begin{small}
\begin{tabular}{|p{0.17\linewidth} | p{0.17\linewidth} | p{0.17\linewidth} | p{0.17\linewidth} | p{0.17\linewidth}|}
\hline
Original & MT (beam) & MT (sampling) & Ours (beam) & Ours (sampling) \\ \hline
The index would fall 3.9\% if the sales of vehicles were not included.&	The index would fall by 3.9 per cent if vehicle sales were not included.&	The Index would fall 3.9\% if the vehicle sales were not included, or 3.9\% if the vehicle sales were excluded.&	The index would decline by 3.9\% if the vehicle sales were not included.&	The index would decline by 3.9\% if vehicles were not included in sales. \\ \hline 
Some people worry that this will affect the business of large-sized Canadian enterprises.&	There are concerns that this may affect the operations of large Canadian companies.&	There is concern that this may affect what large Canadian firms have in their business.&	Some may be concerned that this will affect the major Canadian enterprises.&	Some people worry that this will impact on the great enterprises in Canada. \\ \hline 
These people can set examples and they can have direct influence over the improvement of local human rights conditions and the protection of employees.&	These individuals can provide examples and have a direct influence on the improvement of local human rights and employee protection conditions.&	These may lead to examples and direct influence on the betterment of local human rights conditions and on the protection of wage earners.&	They can provide examples and can directly influence the improvement of local human rights conditions and the protection of employees.&	They can provide examples and have direct influence on improving local human rights and the protection of employees' conditions. \\ \hline
The National Youth League said these activities are aimed at showing support and adoration for the state leaders.&The National Youth League stated that these activities are aimed at showing support and admiration to State leaders.&The National Youth League (NDY) stated that these activities are aimed at demonstrating support and admiration for State leaders.&The National League of Youth indicated that these activities are intended to provide support and encourage state leaders.&The National Youth League has stated that such activities are aimed at showing support and admiration for State leaders.
\\ \hline

\end{tabular}
\end{small}
\caption{Sample paraphrases generated by the different methods.}
\label{tab:examples}
\end{table*}

\subsection{Qualitative analysis}
\label{sec:humaneval}

So as to better understand the behavior of our approach in comparison with round-trip MT, we carried out a human evaluation through Amazon Mechanical Turk. Following the setup of \citet{hosking2021factorising}, we sample 200 sentences from the MTC corpus and generate a pair of paraphrases for each of them, randomly choosing two systems to generate them. We then ask human evaluators to compare the two sentences according to three criteria: diversity, fidelity and fluency. %
More details about the judging criteria can be found in Appendix~\ref{app:judging}.

Figure \ref{fig:human} reports the percentage of head-to-head comparisons that each system has won. %
The results that we obtain are consistent with the trends observed in \S\ref{sec:automaticeval}. More concretely, we observe that the beam search variant of round-trip MT achieves the best results in terms of fluency and fidelity, but does worst in diversity, indicating a tendency to copy. Our method with beam search does slightly better than the sampling MT variant in terms of diversity and slightly worse in terms of fidelity ---indicating a tendency to favor diversity over fidelity--- while also being more fluent. Finally, the sampling variant of our method achieves the best diversity, but has the worst fidelity and fluency.

So as to further contrast these results, we manually analyzed some paraphrases,\footnote{We randomly sampled 20 sentences from MTC and chose four illustrative examples for Table \ref{tab:examples}. The 20 random examples are shown in Appendix \ref{app:samples}.} and report some examples in Table \ref{tab:examples}. Just in line with our previous results, we observe that the beam search variant of round-trip MT tends to deviate the least from the original sentence, while the sampling variant of our method generates the most diverse paraphrases (e.g., changing \textit{``sales  of vehicles  were  not  included''} to \textit{``vehicles were not included in sales''}). At the same time, we observe that this tendency to improve diversity can cause artifacts like paraphrasing named entities (e.g., changing \textit{``National Youth League''} to \textit{``National League of Youth''}), which can partly explain the drop in fidelity.

\section{Conclusions} 

In this work, we have shown that the implicit similarity function present in round-trip MT is not appropriate in the general case, as it considers sentence pairs that share a single ambiguous translation to be paraphrases. We address this issue by designing an alternative similarity function that requires the entire translation distribution to match, and develop a relaxation of it through the IB method, which we prove to be less susceptible to the problem of confounding translations. We implement this approach through adversarial learning, training an encoder to preserve as much information as possible about the reference translation, while encoding as little as possible about the source. Not only is our approach more principled than round-trip MT, but it is also more efficient at inference, as it does not need to generate an intermediate translation. In addition, it offers a knob to adjust the fidelity-diversity trade-off through the $\lambda$ parameter, and obtains strong results in our experiments, outperforming round-trip MT.

\section*{Acknowledgments}

Aitor Ormazabal, Gorka Labaka, Aitor Soroa and Eneko Agirre were supported by the Basque Government (excellence research group IT1343-19 and DeepText project KK-2020/00088) and the Spanish MINECO (project DOMINO PGC2018-102041-B-I00 MCIU/AEI/FEDER, UE). Aitor Ormazabal was supported by a doctoral grant from the Spanish MECD. Computing infrastructure funded by UPV/EHU and Gipuzkoako Foru Aldundia.

\bibliography{acl2022}
\bibliographystyle{acl_natbib}

\appendix

\section{Round-trip MT with restricted sampling}
\label{app:topk}

Our formulation in Section \ref{sec:impfunc} considers all possible translations into the pivot language. In practice, some round-trip MT systems use restricted sampling, considering only a subset of $Y$. For example, ParaNET \citep{mallinson-etal-2017-paraphrasing} takes the $K$ highest probability translations given by beam search. As we show next, the fundamental analysis in Section \ref{sec:impfunc} still holds in that case, and the problem of confounding translations can even be exacerbated by it.

More concretely, using this pivot selection strategy yields the following adjusted paraphrase probability:
\begin{equation}
\begin{split}
P(x_p | x_s) = P(x_p) \sum_{y \in  \{y_1,...,y_K\} }\frac{ P(y | x_s) P(y|x_p)}{ZP(y)} ,
\end{split}
\end{equation}
where $\{y_1, ..., y_K\}$ are the top translation candidates and $Z=  \sum_{y \in \{y_1,...,y_K\} } P(y | x_s)$. In general, if a subset $S(x_s) \in Y$ of the translation space is considered as pivots, the paraphrase probability will be 
\begin{equation}
\begin{split}
P(x_p | x_s) = P(x_p) \sum_{y \in S(x_s) }\frac{ P(y | x_s) P(y|x_p)}{Z(x_s)P(y)} =  \\ \frac{P(x_p)}{Z(x_s)} S'_{MT}(x_p, x_s),
\end{split}
\end{equation}
where $Z(x_s) =  \sum_{y \in S(x_s)} P(y | x_s)$ is a normalizing factor that doesn't depend on the paraphrase $x_p$, and $S'_{MT}(x_p, x_s)$, is the same similarity function as $S_{MT}$, with the sum over $y$ restricted to $S(x_s)$.

Using restricted pivot selection strategies such as beam search, top-K sampling, or nucleus sampling \citep{Holtzman2020The} will yield different pivot subsets $S(x_p)$, which will lead to paraphrase probabilities being assigned based on a limited subset of the entire translation distribution. This can exacerbate the issues outlined in Section \ref{sec:impfunc}, where the similarity metric can be dominated by a single shared high-probability translation. For example, in the case of translating from a gendered to a genderless language, while the highest probability translation will be genderless, a lower probability candidate might identify the gender, so skipping this translation sampling would increase the similarity of sentences that differ only in gender.

\section{Characterizing the encoding learned through the IB method}
\label{sec:charac}
Since we will not learn a perfect minimal sufficient statistic in practice, it is desirable to characterize what the relaxation of $S$ implemented by the IB method can learn.

To that end, we will characterize the kind of encoding $T$ that is allowed by a given $\gamma$.  
Since $T$ is a function of $X$, we know that $I(X,Y)\geq I(T,Y)$, and thus the condition $I(T,Y)\geq \gamma$ can be rewritten as $I(X,Y) - I(T,Y) \leq \epsilon$, where $\epsilon\geq 0$, which is the form we use throughout this section.

Now, for matters of conditional translation probabilities $P(y | T(x))$, the encoding $T$ can be fully characterized by the equivalence relation it defines on $X$, where $x_1\sim x_2$ iff $T(x_1)=T(x_2)$. Two sentences will induce the same conditional translation distribution $P(y | T(x))$ when they are clustered into the same equivalence class by $T$. 
\begin{theorem}

\label{theorem:charac}
Let $\mathcal{S}$ denote the partition of $X$ induced by the encoding $T$. We denote the elements of a cluster $S\in \mathcal{S}$ by $S = \{x^S_1, ... , x^S_{m_S}\}$.  Then, the information loss $I(X,Y) - I(T,Y)$ is given by:

\begin{equation}
\begin{split}
I(X,Y) - I(T,Y) =
   \sum_{S\in \mathcal{S}} P(x_1^S) KL(P_{Y|x_1^S} || P_{Y|S}) \\
     + ...  + P(x_{m_S}^S) KL(P_{Y|x_{m_S}^S} || P_{Y|S}),
\end{split}
\end{equation}

and the translation probabilities conditioned on a cluster are given by the mixture distribution $P(y | T(x)) = P(y | x\in S) = \alpha_1 P(y | x_1^S) + ... + \alpha_{m^S} P(y| x_{m^S}^S)$, where $\alpha_i = \frac{P(x_i^S)}{P(x_1^S) + ... + P(x_{m^S}^S)}$.

\end{theorem}

The proof can be found in Section \ref{app:proofcharac}. This theorem expresses the information loss of an encoding $T$, $I(X,Y)- I(T,Y)$, in terms of the KL divergences between the translation distributions of source sentences $P(y|x)$ and the translation distributions given their encodings $P(y|T(x)) = P(y|x\in S)$. Intuitively, if $T$ clusters together two sentences $x_1$ and $x_2$ (i.e. $T(x_1)=T(x_2)$ holds),  such that $P_{Y|x_1}$ and $P_{Y|x_2}$ are very different, then the mixture distribution $P_{Y| S}$ will be different from both, and thus the information loss will be large. 

We will now obtain more intuitive bounds for the information loss. As seen before, the translation distribution given a cluster $P(y|x\in S)$ can be expressed as a mixture of the individual translation distributions for sentences in the cluster:

\begin{equation}
\begin{split}
P(y|x\in S) = 
\frac{P(x_1^S)}{P(x_1^S) + ... + P(x_{m^S}^S)} P(y|x_1^S) \\ + ... + \frac{P(x_{m^S}^S)}{P(x_1^S) + ... + P(x_{m^S}^S)} P(y|x_{m_S}^S). 
\end{split}
\end{equation}

We can also define mixtures of all the distributions $P(y|x_i^S)$ except one, with the same weights as in $P(y|x\in S)$, except for a re-normalization constant. Explicitly, we define:

\begin{equation}
\begin{split}
P^S_j(y) =  \\
\sum_{i=1, i\neq j }^{m_S} \frac{P(x_i^S)P(y|x_i^S)}{P(x_1^S) + ... + \widehat{P( x_{j}^S)}+... + P(x_{m^S}^S)} ,
\end{split}
\end{equation}

where the hat indicates that that element is skipped. Then, we have the following theorem: 

\begin{theorem}
\label{theorem:l1bound}
Let $\mathcal{S}$ be the partition imposed by the encoding function $T$ on $X$. We denote the elements of a cluster $S\in \mathcal{S}$ by $S = \{x^S_1, ... , x^S_{m_S}\}$. We define the partial mixtures $P^S_j(y)$ as above. Then, if the information loss satisfies  $I(X,Y)- I(T,Y) \leq \epsilon $,  we have

\begin{equation}
\begin{split}
   \sum_{S\in \mathcal{S}} \sum_{i=1}^{m^S} \frac{P(x_i^S)(\beta^S_i)^2}{2} D_1(P_{Y|x_i^S}, P^S_i)^2 \leq \epsilon,
\end{split}
\end{equation}
where $\beta^S_j = \frac{P(x^S_{1}) + ... + \widehat{P( x_{j}^S)} + ... + P(x^S_{m^S}) }{P(x^S_{1}) + ... + P(x^S_{m^S})}$ and $D_1$ is the $L^1$ norm distance.

\end{theorem}

The proof can be found in Section \ref{app:proofl1}. Intuitively, this states that, if the encoding $T$ clusters a set of sentences $x_1,...,x_n\in S$ together, then the translation distribution for an element $x_i\in S$, $P_{Y|x_i}$, cannot be too far from the mixture of the rest of the distributions $P_{Y|x_j}$, with $j\neq i$. 

In the scenario where there are only two sentences in a cluster, $m^S=2$, we have $P_1 = P_{Y|x_2^S}$ and $P_2 = P_{Y|x_1^S}$, and the inner sum reduces as follows (the derivation is shown in Section \ref{sec:derivationbinary}):

\begin{equation}
\label{eq:binarycluster}
\begin{split}
\sum_{i=1}^{m^S} \frac{P(x_i^S)(\beta^S_i)^2}{2} D_1(P_{Y|x_i^S}, P^S_i)^2 
\\= 
  \frac{P(x_1^S)P(x_2^S)}{2(P(x_1^S) + P(x_2^S))}   D_1(P_{Y|x_1^S}, P_{Y|x_2^S})^2
\end{split}
\end{equation}

Since clustering all the elements of a set $S$ leads to a bigger information loss than only clustering any two elements $x_1,x_2\in S$, combining Equation $\ref{eq:binarycluster}$ and Theorem $\ref{theorem:l1bound}$ we obtain Theorem \ref{theorem:7} from \S\ref{subsec:ib-characterization} as a corollary:

\begingroup
\def\thetheorem{\ref{theorem:7}}
\begin{theorem}

Suppose $T$ is a solution of the IB optimization problem $\min_T I(X,T) \; s.t \; I(T,Y) \geq \gamma$, and $\epsilon = I(X,Y) - \gamma$. If $\mathcal{S}$ is the partition on $X$ induced by $T$, we have: 
\begin{equation}
\begin{split}
\sum_{S\in \mathcal{S}} \max_{x_1, x_2\in S} \frac{P(x_1)P(x_2)}{2(P(x_1) + P(x_2))}  D_1(P_{Y|x_1}, P_{Y|x_2})^2 \\\leq \epsilon,
\end{split}
\end{equation}
where $D_1$ is the $L_1$ norm distance.

\end{theorem}
\addtocounter{theorem}{-1}
\endgroup

This bound is the easiest to interpret intuitively, as it bounds the pairwise distances between the translation distributions of any two sentences that are considered paraphrases by the encoding $T$. To sum up the results from this section, the information loss allowance when learning with the IB method bounds the $L^1$ norm distance between the translation distributions of paraphrases. Thus the entire translation distribution is considered when learning paraphrases, potentially alleviating the problems discussed in Section \ref{sec:impfunc}

\section{Proofs}
\label{app:proofs}
\subsection{Proof of Theorem \ref{theorem:IB}}
\label{app:proofminsuff}
We know that $T$ is a minimal sufficient statistic of $Y$ if and only if the following condition is satisfied:
\begin{equation}
    \begin{split}
    \frac{P(x|y)}{P(x'|y)} \text{ independent of y} \iff \\ T(x) = T(x') \; \forall x, x' \in X 
    \end{split}
\end{equation}

Rewriting $ P(x|y) = \frac{P(y|x)P(x)}{P(y)}$ and cancelling terms, the $LHS$ becomes:
\begin{equation}
    \frac{P_{y}(x)}{P_{y}(x')} = \frac{P(y|x)}{P(y|x')}\frac{P(x)}{P(x')}.
\end{equation}

Since $\frac{P(x)}{P(x')}$ does not depend on $y$, the entire term will not depend on $y$ if and only if $\frac{P(y|x)}{P(y|x')}$ is independent of $y$. It is easy to see that the ratio of two distributions of $y$ will be independent of $y$ if and only if they are the exact same distribution, and thus we can conclude that if $T$ is a minimal sufficient statistic of $Y$ then $T(x) = T(x') \iff P(y|x) = P(y|x') \forall y\in Y $, or, equivalently, $T(x) = T(x') \iff S(x, x') = 1$. 

Thus, we have 
\begin{equation}
\begin{split}
    P&(x_p|T(x_s)=P(X=x_p | T(X) = T(x_s))
    \\&= P(X=x_p | S(X,x_s) = 1) 
    \\&=\frac{P(X=x_p,  S(X,x_s) = 1)}{P(S(X,x_s)=1)} 
    \\&= 
    \frac{P(X=x_p) P( S(X,x_s) = 1)| X=x_p)}{P(S(X,x_s)=1)}
    \\&=  \frac{P(x_p) S(x_p,x_s) }{Z},
\end{split}
\end{equation}

where $Z = P(S(X,x_s)=1)$ is the normalizer that does not depend on $x_p$, as we wanted to prove. \qed

\subsection{Proof of Theorem \ref{theorem:charac}}
\label{app:proofcharac}

We first expand the information loss:

\begin{equation}
\begin{split}
I(X,&Y) - I(T,Y)
\\&= \mathbb{E}_X KL(P_{Y|X} || P_Y) 
\\&-  \mathbb{E}_T KL(P_{Y|T} || P_Y)
= \sum_x P(x)  
\\ &\bigg[\sum_y P(y|x)log(P(y|x)) 
\\ &- P(y|x)log(P(y))\bigg]
\\&- \sum_x P(x) \bigg[\sum_y P(y|T(x))log(P(y|T(x)))
\\&-  P(y|T(x))log(P(y))\bigg]
\end{split}
\end{equation}

Now, for matters of conditional translation probabilities $P(y | T(x))$, the encoding $T$ can be fully characterized by the equivalence class it defines on $X$, where $x_1\sim x_2$ iff $T(x_1)=T(x_2)$. We let $\mathcal{S}$ denote the partition on $X$ induced by this equivalence class. Then, we can rewrite:

\begin{equation}
\label{eq:first}
\begin{split}
 &I(X,Y) - I(T,Y) 
 \\&= \sum_x P(x)\bigg[ \sum_y P(y|x)log(P(y|x))
 \\&-\ P(y|x)log(P(y)) \bigg]
 \\&- \sum_x P(x) \bigg[\sum_y P(y|T(x))log(P(y|T(x)))
 \\&- P(y|T(x))log(P(y)) \bigg]
 \\& =  \sum_{S\in \mathcal{S}} \sum_{x\in S} P(x) \bigg[\sum_y P(y|x)log(P(y|x))
 \\&-P(y|x)log(P(y))  \bigg]
 \\&-\sum_{S\in \mathcal{S}} \sum_{x\in S}  P(x) \bigg[\sum_y P(y|x\in S)log(P(y|x\in S)) 
 \\&-P(y|x\in S)log(P(y)) \bigg]
\end{split}
\end{equation}

For a certain $S\in \mathcal{S}$, we denote its elements by $S = \{x^S_1, ... , x^S_{m_S}\}$. Then, we have 

\begin{equation}
\begin{split}
\label{eq:mixtureexpression}
P(y|x\in S) &= \frac{P(y, x\in S)}{P(x\in S)} 
\\&=
\frac{P(y, x_1^S) + ... + P(y, x_{m_S}^S)}{P( x_1^S) + ... + P(x_{m_S}^S)} 
\\&=
\alpha^S_1 P(y|x_1^S) + ... +\alpha^S_{m_S} P(y|x_{m_S}^S),
\end{split}
\end{equation}

where $\alpha_i^S = \frac{P(x^S_i)}{P( x_1^S) + ... + P(x_{m_S}^S)}$. We also define $\beta^S = P( x_1^S) + ... + P(x_{m_S}^S) $. Now, we can rewrite the first expression of the RHS in Equation \ref{eq:first}:

\begin{equation}
\label{eq:firstterm}
\begin{split}
 &\sum_{S\in \mathcal{S}} \sum_{x\in S} P(x) \sum_y \big[ P(y|x)log(P(y|x)) 
 \\&- P(y|x)log(P(y)) \big] 
 \\&=  \sum_{S\in \mathcal{S}} \bigg[ P(x^S_1) \sum_y P(y|x^S_1)log(P(y|x^S_1))+ ... 
 \\&+ P(x^S_{m_S}) \sum_y  P(y|x^S_{m_S})log(P(y|x^S_{m_S}))   \bigg]    
 \\&- \bigg[\sum_y  P(x^S_1)P(y|x^S_1)log(P(y)) + ... \\&+  P(x^S_{m_S})P(y|x^S_{m_S})log(P(y)) \bigg] 
 \\&= \sum_{S\in \mathcal{S}} \bigg[ P(x^S_1) \sum_y P(y|x^S_1)log(P(y|x^S_1))+ ... 
 \\&+ P(x^S_{m_S}) \sum_y  P(y|x^S_{m_S})log(P(y|x^S_{m_S}))   \bigg]  
 \\&- \bigg[\sum_y  \beta^S \alpha^S_1 P(y|x_1^S)log(P(y)) + ...
 \\& +  \beta^S \alpha^S_{m_S}P(y|x_{m_S}^S)log(P(y)) \bigg] 
 \\&=\sum_{S\in \mathcal{S}} \bigg[ P(x^S_1) \sum_y P(y|x^S_1)log(P(y|x^S_1))+ ... 
 \\&+ P(x^S_{m_S}) \sum_y  P(y|x^S_{m_S})log(P(y|x^S_{m_S}))   \bigg]    
 \\&- \bigg[\beta^S \sum_y P(y|x\in S) log(P(y)) \bigg]
\end{split}
\end{equation}

And now we rewrite the second term in the RHS of Equation \ref{eq:first}:

\begin{equation}
\label{eq:secondterm}
\begin{split}
 &\sum_{S\in \mathcal{S}} \sum_{x\in S} P(x) \sum_y \big[ P(y|x\in S)log(P(y|x\in S)) 
 \\&- P(y|x\in S)log(P(y)) \big] 
 \\&= \sum_{S\in \mathcal{S}} \bigg[ P(x^S_1) \sum_y   P(y|x\in S)log(P(y|x\in S)) 
 \\& + P(x^S_{m_S}) \sum_y   P(y|x\in S)log(P(y|x\in S)) \bigg] 
 \\&- \bigg[P(x^S_1)\sum_y  P(y|x\in S)log(P(y)) 
 \\& + P(x^S_{m_S})\sum_y  P(y|x\in S)log(P(y))\bigg]  
 \\&= \sum_{S\in \mathcal{S}} \bigg[  \beta^S \sum_y   P(y|x\in S)log(P(y|x\in S)) \bigg] 
 \\&- \bigg[\beta^S\sum_y  P(y|x\in S)log(P(y))\bigg] 
 \\&= \sum_{S\in \mathcal{S}} \bigg[ P(x^S_1) \sum_y   P(y|x^S_1)log(P(y|x\in S)) 
\\&+ P(x^S_{m_S}) \sum_y   P(y|x^S_{m_S})log(P(y|x\in S)) \bigg] 
 \\&- \bigg[\beta^S\sum_y  P(y|x\in S)log(P(y))\bigg], 
\end{split}
\end{equation}

where we have used Equation \ref{eq:mixtureexpression} in the last equality.

Substituting (\ref{eq:firstterm}) and (\ref{eq:secondterm}) into (\ref{eq:first}), we get:

\begin{equation}
\begin{split}
&I(X,Y) - I(T,Y) 
\\&=\sum_{S\in \mathcal{S}} \bigg[ P(x^S_1) \sum_y P(y|x^S_1)log(P(y|x^S_1))+ ...
\\ &+ P(x^S_{m_S}) \sum_y  P(y|x^S_{m_S})log(P(y|x^S_{m_S}))   \bigg] 
\\&- \sum_{S\in \mathcal{S}} \bigg[ P(x^S_1) \sum_y   P(y|x^S_1)log(P(y|x\in S)) 
\\&+ P(x^S_{m_S}) \sum_y   P(y|x^S_{m_S})log(P(y|x\in S)) \bigg] 
\\&= \sum_{S\in \mathcal{S}} P(x_1^S) \sum_y \bigg[ P(y|x^S_1)log(P(y|x^S_1)) 
\\& - P(y|x^S_1)log(P(y|x\in S)) \bigg] + ... 
\\&+ P(x_{m_S}^S) \sum_y \bigg[ P(y|x_{m_S}^S)log(P(y|x_{m_S}))
\\& - P(y|x_{m_S}^S)log(P(y|x\in S)) \bigg] 
\\&= \sum_{S\in \mathcal{S}} P(x_1^S) KL(P_{Y|x_1^S} || P_{Y|S})  + ... 
\\& + P(x_{m_S}^S) KL(P_{Y|x_{m_S}^S} || P_{Y|S})
\end{split}
\end{equation}

As we wanted to show. \qed 

\subsection{Proof of Theorem \ref{theorem:l1bound}}
\label{app:proofl1}
By Theorem \ref{theorem:charac}, it is enough to show that 

\begin{equation}
\begin{split}
   \sum_{S\in \mathcal{S}} P&(x_1^S) KL(P_{Y|x_1^S} || P_{Y|S}) \\
    & + ...  + P(x_{m_S}^S) KL(P_{Y|x_{m_S}^S} || P_{Y|S}) \\&\geq  \sum_{S\in \mathcal{S}} \sum_{i=1}^{m^S} \frac{P(x_i^S)(\beta^S_i)^2}{2} D_1(P_{Y|x_i^S}, P^S_i)^2 
\end{split}
\end{equation}

For that, it is enough to show that 
\begin{equation}
\begin{split}
   P(x_i^S)& KL(P_{Y|x_i^S} || P_{Y|S}) 
   \\  &\geq   \frac{P(x_i^S)(\beta^S_i)^2}{2} D_1(P_{Y|x_i^S}, P^S_i)^2 
\end{split}
\end{equation}

for every i. Now, by Pinsker's inequality, for a given i, we have that 

\begin{equation}
    \begin{split}
        P&(x_i^S) KL(P_{Y|x_i^S} || P_{Y|S}) =P(x_i^S) 
        \\&KL(P_{Y|x_i^S} ||  \alpha_1 P_{Y | x_1^S} + ... + \alpha_{m^S} P_{Y| x_{m^S}^S} )
        \\&\geq \frac{P(x_i^S)}{2}D_1(\alpha_1 P_{Y | x_1^S} + ... 
        \\&+ \alpha_{m^S} P_{Y| x_{m^S}^S} , P_{Y | x_i^S})^2
        \\& = \frac{P(x_i^S)}{2( P(x_1^S) + ... + P(x_{m^S}^S) )^2}||_1(P(x_1) P_{Y | x_1^S} + ...
         \\ &+ P(x_{m^S}) P_{Y| x_{m^S}^S} 
         \\&- ( P(x_1^S) + ... + P(x_{m^S}^S)) P_{Y | x_i^S}||^2
        \\& =\frac{P(x_i^S)}{2( P(x_1^S) + ... + P(x_{m^S}^S) )^2}
        \\&||_1(P(x_1^S) P_{Y | x_1^S} + ... \widehat{P(x_i^S)P_{Y | x_i^S}}+ ...
        \\&+ P(x_{m^S}^S) P_{Y| x_{m^S}^S} 
        \\&- ( P(x_1^S) + ... + \widehat{P(x_i^S)}+ ... + P(x_{m^S})) P_{Y | x_i^S}||^2
        \\& = \frac{P(x_i^S)( P(x_1^S) + ... + \widehat{P(x_i^S)}+  ... + P(x_{m^S}) )^2}{2( P(x_1^S) + ... + P(x_{m^S}^S) )^2}
        \\&||_1P_i^S -  P_{Y | x_i^S}||^2
        \\&= \frac{P(x_i^S)(\beta _i^S)^2}{2} D_1(P_i^S(Y), P_{Y | x_i^S})^2
    \end{split}
\end{equation}

where the hat represents that element of the sum being skipped, as we wanted to show. \qed 

\subsection{Derivation of Equation \ref{eq:binarycluster} }
\label{sec:derivationbinary}

\begin{equation}
\begin{split}
    &\sum_{i=1}^{m^S} \frac{P(x_i^S)(\beta^S_i)^2}{2} D_1(P_{Y|x_i^S}, P^S_i)^2  
    \\& = \bigg[  \frac{P(x_1^S)(\beta^S_1)^2}{2} +  \frac{P(x_2^S)(\beta^S_2)^2}{2} \bigg] D_1(P_{Y|x_1^S}, P_{Y|x_2^S})^2
    \\&= \bigg[  \frac{P(x_1^S)P(x_2^S)^2}{2(P(x_1^S) + P(x_2^S))^2} +  \frac{P(x_2^S)P(x_1^S)^2}{2(P(x_1^S) + P(x_2^S))^2} \bigg]
    \\& D_1(P_{Y|x_1^S}, P_{Y|x_2^S})^2 
    \\& =  \bigg[  \frac{P(x_1^S)P(x_2^S)(P(x_1^S) + P(x_2^S)) }{2(P(x_1^S) + P(x_2^S))^2}  \bigg]
\\ &D_1(P_{Y|x_1^S}, P_{Y|x_2^S})^2
\\&=  \frac{P(x_1^S)P(x_2^S)}{2(P(x_1^S) + P(x_2^S))}   D_1(P_{Y|x_1^S}, P_{Y|x_2^S})^2
\end{split}
\end{equation}

\section{Human evaluation judging criteria}
\label{app:judging}

We ask human evaluators to compare systems on three different dimensions, according to the following instructions:

\textbf{Meaning. }Which of the paraphrases better preserves the meaning of the original, without adding or losing information?

\textbf{Surface similarity. } Which of the paraphrases is more similar compared to the original, using more similar phrasing or words? You should chose the text using more similar words or phrasing, regardless of meaning.

\textbf{Fluency.} Which text is a more fluent English sentence? You should choose the sentence that contains the least grammatical mistakes, and sounds more natural.

\section{Full paraphrase sample}
\label{app:samples}
We present the full list of 20 sampled paraphrases in Table \ref{tab:fullexamples}.

\begin{table*}
\begin{small}
\begin{tabular}{|p{0.17\linewidth} | p{0.17\linewidth} | p{0.17\linewidth} | p{0.17\linewidth} | p{0.17\linewidth}|}
 \hline 

Original & MT (beam) & MT (sampling) & Ours (beam) & Ours (sampling) \\ \hline
As of August 30th, the city had allocated a labor force of 158,300 per day in the project, with the aggregate labor contribution amounting to 936,500, and had completed 730,000 cubic meters of earth and stone. & As at 30 August, the city had allocated 158,300 persons per day to the project, with an overall contribution of 936,500 persons, and had completed the construction of 730,000 cubic metres of land and stone. & It had assigned 158,300 employees per day to the project as of 30 August, with a total staff contribution of 936,500, and it had concluded the construction of 730,000 cubic metres of earth and rock. & As of 30 August, the city had employed 158,300 persons per day for the project, a total of 936,500 of whom had been allocated for the construction of 230,000 square meters of earth and stone. & On 30 August, the City had a staff of 158,300 people per day in the project, a total of 936,500 of whom had been provided for the construction of 330,000 square foot of earth andstone work.\\ \hline
The United States and North Korea are scheduled to hold talks on Friday regarding US access to a suspected underground nuclear site in the vicinity of Pyongyang. & The United States and North Korea are expected to hold talks on Friday on United States access to a suspected underground nuclear site near Pyongyang. & The United States and North Korea are scheduled for talks Friday on the US access to a suspected underground nuclear site near Pyongyang. & The United States and North Korea are scheduled to hold talks Friday on the access of the United States to a suspected nearby Pyongyang ground nuclear site. & The United States and North Korea are scheduled to hold talks on Friday on United States access to a suspected nearby Pyongyang nuclear site.\\ \hline
For long, Xi'ning has made insufficient investment in the construction of urban infrastructure facilities, with the total investment made being only about RMB400 million during the 46-year period between the founding of the New China to 1995. As a result, the backward and underdeveloped infrastructure facilities have restricted the city's economic development. & Xi'ning has for a long time not invested sufficiently in the construction of urban infrastructure, with a total investment of only about RMB 400 million over the 46-year period between the founding of New China and 1995, as a result of which lagging and underdeveloped infrastructure has limited the city's economic development. & Xi'ning had for a long time not invested enough in construction of urban infrastructure, with total investment amounting only to around RMB 400m during the 46-year period between the founding of New China and 1995. Therefore, backward and underdeveloped infrastructure had impeded economic development of the city. & Since long Xi'ning's investment in urban infrastructure has been insufficient, with a total investment of only 400 million cubic metres in total, for 46 years between the founding of the New China in 1995, thus limiting the development of infrastructure in the underdeveloped and underdeveloped areas. & Since long Xi'ning has not made adequate investment in urban infrastructure, with an overall investment of only \$400 million MB over 46 years from the creation of the new China to 1995, thus restricting the development of urban infrastructure, both underdeveloped and underdeveloped.\\ \hline
Japan, Australia, New Zealand and South Korea Expresses Support, saying that the U.S. Has No Other Choice & Japan, Australia, New Zealand and South Korea expressed their support, stating that the United States had no other choice. & Japan, Australia, New Zealand and South Korea show their support, affirming that the United States does not have no Alternative & Japan, Australia, New Zealand and South Korea expressed their support, stating that the United States had no other choice. & Japan, Australia, New Zealand and South Korea expressed support, stating that the United States had no other choice.\\ \hline
Chernomyrdin also pointed out that there were also many problems in Russia last year, such as the poor performance of taxation, investment plans yet to be completed, reduced surplus in foreign trade, and ineffective fiscal and financial measures taken by the government. & Chernomyrdin also pointed out that there were also many problems in Russia last year, such as poor fiscal performance, investment plans that had not yet been completed, the reduction of the foreign trade surplus and ineffective fiscal and financial measures taken by the government. & Chernomyrdin also stressed that, in Russia, last year, there were many problems too. These included a deteriorating fiscal performance, investment plans that have yet to be completed, the reduction of foreign trade surpluses, and inefficient fiscal and financial measures taken by the Government, etc. & Chernomyrdin also noted that there were many problems in Russia last year, such as poor taxation performance, plans for investment still to be completed, the reduction in foreign trade surplus, and inadequate fiscal and financial measures taken by the Government. & Chernomyrdin further noted that there were many further problems in Russia last year, such as weak taxation performance, plans for further investment still to be completed, the reduction in foreign trade surplus, and ineffective fiscal and financial measures taken by the Government.\\ \hline
Based on the plan, the GDP in Russia next year is to increase by 2\%, and the inflation rate is to go down to 5\% to 8\%. & According to the plan, Russia's GDP will rise by 2 per cent next year and the inflation rate will rise from 5 per cent to 8 per cent. & According to the plan, Russian GDP will rise 2 per cent next year and the rate of inflation will be reduced from 5 per cent to 8 per cent. & According to this plan, GDP in Russia will increase by 2\% next year and the inflation rate will drop from 5\% to 8\%. & Under the plan, the GDP in Russia will increase 2 per cent next year and the uninflation rate increase from 5\% to 8\%.\\ \hline

\end{tabular}
\end{small}
\end{table*}

\begin{table*}
\begin{small}
\begin{tabular}{|p{0.17\linewidth} | p{0.17\linewidth} | p{0.17\linewidth} | p{0.17\linewidth} | p{0.17\linewidth}|}
\hline
Zuo Zhongmo, deputy secretary-general of the Conference, said, "This is not just an issue of agriculture. These reclaimed lands can serve the general development purposes of various sectors, including forestry, industry and tourism." & Zuo Zhongmo, Under-Secretary-General of the Conference, said: “This is not just an agriculture issue, these recovered lands can serve the overall development goals of various sectors, including forestry, industry and tourism. & Zuo Zhongmo, Deputy Secretary-General of the Conference, said, "It is not just an agriculture issue; the land recovered from them can serve the overall development goals of different sectors including forestry, industry and tourism." & It is Zuo Zhongmo, Under-Secretary-General of the Conference, who said: It is not just about agriculture; the reclaimed lands can be used for general development in various sectors, including forestry, industry and tourism." & The Assistant Secretary-General of the Conference, Zuo Zhongmo, said, It is not just about agriculture, as the reclaimed lands can be used for general development purposes from various sectors, including forestry, industry and tourism.\\ \hline
In the United States, California and other southern states were flooded at the beginning of this year, followed by a drought in many places in the south. & In the United States, California and other southern states were flooded early this year, followed by droughts in many parts of the South. & In the United States of America, CaliFORNA and other southern states were flooded early this year, followed by drought in many parts of the south. & In the United States, California and other southern states, floods occurred in early this year, followed by drought in many areas of the south. & Inondations in California, and other southern States in early this year, followed by droughts in many endroits in the south.\\ \hline
Meanwhile, the US Congress is discussing whether or not to approve the Protocol reached in the Kyoto Conference in Japan. & In the meantime, the United States Congress is discussing whether or not to approve the Protocol concluded at the Kyoto Conference in Japan. & Meanwhile, US Congress debates whether or not to accede to the Protocol agreed at the Kyoto Conference in Japan. & At the same time, the United States Congress is examining whether or not to approve the Protocol at the Kyoto Conference in Japan. & At the same time, the United States Congress is considering whether or not to approve the Protocol made at the Kyoto Conference in Japan.\\ \hline
The index would fall 3.9\% if the sales of vehicles were not included. & The index would fall by 3.9 per cent if vehicle sales were not included. & The Index would fall 3.9\% if the vehicle sales were not included, or 3.9\% if the vehicle sales were excluded. & The index would decline by 3.9\% if the vehicle sales were not included. & The index would decline by 3.9\% if vehicles were not included in sales.\\ \hline
According to the company, in the coming five years, the company will make an additional investment of US\$90 million, with an anticipated annual output value of US\$300 million. & According to the company, over the next five years, it will make an additional investment of US\$90 million, with a projected annual production value of US\$300 million. & According to the company, it will make an additional \$90 million US over the next five years with a planned annual productivity value of \$300 million US. & According to the company, over the next five years it will invest an additional \$90 million in the U.S., its expected annual output of \$300 million. & According to the company, it will invest \$90 million in the next five years, with its expected annual output of approximately \$300 million.\\ \hline
Some people worry that this will affect the business of large-sized Canadian enterprises. & There are concerns that this may affect the operations of large Canadian companies. & There is concern that this may affect what large Canadian firms have in their business. & Some may be concerned that this will affect the major Canadian enterprises. & Some people worry that this will impact on the great enterprises in Canada.\\ \hline
A Majority of Hong Kong Residents Decline to Consider Themselves as Chinese & A majority of Hong Kong residents decline to regard themselves as Chinese & The Hong Kong City of Hong Kong’s minority people are not recognising themselves as Chinese & The vast majority of Hong Kong residents feel they are being referred as Chinese & The vast majority of Hong Kong residents have become disenfranchised as Chinese\\ \hline
He said the 83-year-old woman has been hospitalized for over-shock. & He said that the 83-year-old woman had been hospitalized for overheating. & He said that the 83-year-old woman had been hospitalized for overheat. & He stated that a 83-year-old woman had been hospitalized for a headache. & He indicated that a 83-year-old woman had been hospitalized for a head injuries.\\ \hline
However, statistics released by the Immigration Bureau showed that although there were 11,978 new British immigrants coming to Australia between 1996 and 1997, 3,737 people left the country during the same period. & However, statistics published by the Immigration Bureau show that, although there were 11,978 new British immigrants to Australia between 1996 and 1997, 3,737 people left the country during the same period. & However, Immigration Bureau statistics show that although there were 11,978 new British immigrants to Australia from 1996 to 1997, 3,737 individuals left the country during that period. & However, the statistics of the Office of Immigration show that, although 11,978 new British immigrants arrived in Australia between 1996 and 1997, 3,737 had left the country during the same period. & However, the figures from the Immigration Bureau show that, although 11,978 new British immigrants had entered Australia between 1996 and 1997, 3,737 had left the country during the same period.\\ \hline

\end{tabular}
\end{small}
\end{table*}

\begin{table*}
\begin{small}
\begin{tabular}{|p{0.17\linewidth} | p{0.17\linewidth} | p{0.17\linewidth} | p{0.17\linewidth} | p{0.17\linewidth}|}
\hline
(Reuters report from Tokyo)Japanese Finance Minister Kiichi Miyazawa was pressured not to quit office yesterday. & Japan's Finance Minister, Kiichi Miyazawa, was forced yesterday not to leave his post. & The finance minister in Japan, Kiichi Miyazawa, was forced to leave the post yesterday. & ( Tokyo report) Japan's Minister of Finance Kiichi Mi Theawa was pressured not to quit yesterday. & ( Tokyo report) The Japanese Finance Minister Kiichi Mi ichiawa was pressured on not to leave yesterday.\\ \hline
US Admitted Hundreds of Deaths of Iraqi Civilians in Air Strike and UK Reporter Claimed the Target Being Definitely Non-military & US Admitted Hundreds of Deaths of Iraqi Civilians in Air Strike and UK Reporter Claimed the Target Being Definitely Non-military & U.S. Admitted Hundreds of Deaths of Iraqi Civilians in Air Strike and UK Reporter Claimed the Target Being Definitely Non- Military & The US killed hundreds of Iraqi civilians in an air strike, and a reporter in the UK said the target was essentially non-military. & Several hundred dead of Iraqi civilians in air strikes; a UK reporter said that its specific goal is, essentially, non-military.\\ \hline
During the Eighth Five-Year Plan Period (from 1991 to 1995), township enterprises in Fujian Province contributed an aggregate total of RMB18.56 billion in tax, and achieved a total of RMB105.5 billion worth of export commodities. & During the eighth period of the five-year plan (1991-1995), municipal enterprises in Fujian Province paid taxes totalling RMB 18.56 billion and exported products totalling RMB 105.5 billion. & During the eighth period of the Five-Year Plan (from 1991 to 1995), Fujian Provincial Municipal Enterprises had paid taxes totalling RMB 18.56 billion and exported revenues totalling RMB 105.5 billion. & During the eighth five-year plan (from 1991 to 1995), businesses in the townships of Fujian Province contributed a total of \$18.56 billion in tax contributions, \$105.5 billion in commodities for export. & During the eighth five-year plan (from 1991 to 1995), businesses in the townships of Fujian Province contributed tax contributions totalling \$118.56 billion in 1991-95, as well as \$1005.5 billion in commodities for export.\\ \hline
In May this year, Dole admitted using Viagra on a trial basis, and gave high remarks to the drug after use, describing it as "a magic drug." & In May of this year, Dole admitted to using Viagra on an experimental basis and commented very positively on the drug after its use, describing it as "a magical drug." & In May this year, Dole admitted that he uses Viagra as an experimental patient and provided very positive feedback on the drug after its use describing it "a magical medicine". & In May of this year, Dole accepted the use of marijuana for trials and reported strong post-treatment remarks, describing it as a magical drug. & In May of this year, Dole recognized the use of pesticide in trials and had reported very good after-treatment, describing it as the “ magical drug”.\\ \hline
These people can set examples and they can have direct influence over the improvement of local human rights conditions and the protection of employees. & These individuals can provide examples and have a direct influence on the improvement of local human rights and employee protection conditions. & These may lead to examples and direct influence on the betterment of local human rights conditions and on the protection of wage earners. & They can provide examples and can directly influence the improvement of local human rights conditions and the protection of employees. & They can provide examples and have direct influence on improving local human rights and the protection of employees' conditions.
\\ \hline

\end{tabular}
\end{small}

 \caption{Sample paraphrases generated by the different methods. }
\label{tab:fullexamples}
\end{table*}

\end{document}